\documentclass{article}

\usepackage[preprint]{nips_2018}

\usepackage{pslatex}

\usepackage[pdftex]{graphicx}
\usepackage{comment}
\usepackage{amsmath}
\usepackage{amssymb}
\usepackage{xcolor}
\usepackage{hyperref}
\usepackage{multirow}
\usepackage{subcaption} 

\title{Further advantages of data augmentation \\ on convolutional neural networks}
 
\author{{\large \bf Alex Hern\'andez-Garc\'ia (ahernandez@uos.de)} \\
  Institute of Cognitive Science, University of Osnabr\"uck\\
  27 Wachsbleiche, 49090 Osnabr\"uck, Germany
  \AND {\large \bf Peter K\"onig\textsuperscript{*} (pkoenig@uos.de)} \\
  Institute of Cognitive Science, University of Osnabr\"uck\\
  27 Wachsbleiche, 49090 Osnabr\"uck, Germany}

\begin{document}

\maketitle

\begin{abstract}
Data augmentation is a popular technique largely used to enhance the training of convolutional neural networks. Although many of its benefits are well known by deep learning researchers and practitioners, its implicit regularization effects, as compared to popular explicit regularization techniques, such as weight decay and dropout, remain largely unstudied. As a matter of fact, convolutional neural networks for image object classification are typically trained with both data augmentation and explicit regularization, assuming the benefits of all techniques are complementary. In this paper, we systematically analyze these techniques through ablation studies of different network architectures trained with different amounts of training data. Our results unveil a largely ignored advantage of data augmentation: networks trained with just data augmentation more easily adapt to different architectures and amount of training data, as opposed to weight decay and dropout, which require specific fine-tuning of their hyperparameters.
\end{abstract}

\begin{quote}
\end{quote}

\section{Introduction}
\label{sec:intro}

Data augmentation in machine learning refers to the techniques that synthetically expand a data set by applying transformations on the existing examples, thus augmenting the amount of available training data. Although the new data points are not independent and identically distributed, data augmentation implicitly regularizes the models and improves generalization, as established by statistical learning theory \cite{vapnik1971vc}.

Data augmentation has been long used in machine learning \cite{simard1992daug} and it has been identified as a critical component of many models \cite{ciresan2011daug,krizhevsky2012alexnet,lecun2015nature}. Nonetheless, the literature lacks, to our knowledge, a systematic analysis of the implicit regularization effect of data augmentation on deep neural networks compared to the most popular regularization techniques, such as weight decay \cite{hanson1989wd} and dropout \cite{srivastava2014dropout}, which are typically used all together. 

In a thought-provoking paper \cite{zhang2016}, Zhang \textit{et al.} concluded that \textit{explicit regularization may improve generalization performance, but is neither necessary nor by itself sufficient for controlling generalization error}.  They observed that removing weight decay and dropout does not prevent the models from generalizing. Although they performed some ablation studies with data augmentation, they considered it just another explicit regularization technique. In a follow up study \cite{hernandez2018weightdecay_dropout}, it is argued that data augmentation should not be considered an explicit regularizer and it is shown that explicit regularization may not only be unnecessary, but data augmentation alone can achieve the same level of generalization.

Here, we build upon the ideas from \cite{hernandez2018weightdecay_dropout} and, using the same methodology, we extend the analysis of data augmentation in contrast to weight decay and dropout. In particular, we focus here on the capability of data augmentation to adapt to deeper and shallower architectures as well as to successfully learn from fewer examples. We find that networks trained with data augmentation, but no explicit regularizers, outperform the networks trained with all techniques, as is common practice in the literature. We hypothesize that weight decay and dropout require fine-tuning of their hyperparameters in order to adapt to new architectures and amount of training data, whereas the new samples generated by data augmentation schemes are useful regardless of the new training conditions.

\subsection{Related work}
\label{sec:relatedwork}

Data augmentation was already used in the late 80's and early 90's for handwritten digit recognition \cite{simard1992daug} and it has been identified as a very important element of many modern successful models, like AlexNet \cite{krizhevsky2012alexnet}, All-CNN \cite{springenberg2014allcnn} or ResNet \cite{he2016resnet}, for instance. In some cases, heavy data augmentation has been applied with successful results \cite{wu2015deepimage}. In domains other than computer vision, data augmentation has also been proven effective, for example in speech recognition \cite{jaitly2013daugspeech}, music source separation \cite{uhlich2017daugmusic} or text categorization \cite{lu2006daugtext}.

Bengio \textit{et al.} \cite{bengio2011daug} focused on the importance of data augmentation for recognizing handwritten digits through greedy layer-wise unsupervised pre-training \cite{bengio2007greedy}. Their main conclusion was that deeper architectures benefit more from data augmentation than shallow networks. Zhang \textit{et al.} \cite{zhang2016} included data augmentation in their analysis of the role of regularization in the generalization of deep networks, although it was considered an explicit regularizer similar to weight decay and dropout. The observation that data augmentation alone outperforms explicitly regularized models for few-shot learning was also made by Hilliard \textit{et al.} in \cite{hilliard2018fewshot}. Only	 few works reported the performance of their models when trained with different types of data augmentation levels, as is the case of \cite{graham2014fracmaxpool}. 

Recently, the deep learning community seems to have become more aware of the importance of data augmentation. New techniques have been proposed \cite{devries2017cutout,devries2017daugfeatspace} and, very interestingly, models that automatically learn useful data transformations have also been published lately \cite{hauberg2016learningdaug,lemley2017smartdaug,ratner2017learningdaug,antoniou2017dagan}. Another study \cite{perez2017dauganalysis} analyzed the performance of different data augmentation techniques for object recognition and concluded that one of the most successful techniques so far is the traditional transformations carried out in most studies. Finally, a preliminary analysis of the implicit regularization effect of data augmentation was presented in \cite{hernandez2018weightdecay_dropout}, showing that data augmentation alone provides at least the same generalization performance as weight decay and dropout. The present work follows up on those results and extends the analysis.

\section{Experimental setup}

This section describes the procedures we follow to explore the potential advantages of data augmentation to adapt to changes in the amount of training data and the network architecture, compared to the popular explicit regularizers weight decay and dropout. We build upon the methodology already used in \cite{hernandez2018weightdecay_dropout}. 

\subsection{Network architectures}

We test our hypotheses with two well-known network architectures that achieve successful results in image object recognition: the all convolutional network, All-CNN \cite{springenberg2014allcnn}; and the wide residual network, WRN \cite{zagoruyko2016wrn}.

\subsubsection{All convolutional net.}
The original architecture of All-CNN consists of 12 convolutional layers and has about 1.3 M parameters. In our experiments to compare data augmentation and explicit regularization in terms of adaptability to changes in the architecture, we also test a \textit{shallower} version, with 9 layers and 374 K parameters, and a \textit{deeper} version, with 15 layers and 2.4 M parameters. The three architectures can be described as follows:

\begin{table}[ht]
\label{tab:allcnn}
\begin{center}
\begin{tabular}{l|c}
\multirow{2}{*}{Original}  & ~2$\times$96C3(1)--96C3(2)--2$\times$192C3(1)--192C3(2)--192C3(1)--192C1(1) \\
                           & --\textit{N.Cl}.C1(1)--Gl.Avg.--Softmax \\ [3pt]
\multirow{1}{*}{Shallower} & ~2$\times$96C3(1)--96C3(2)--192C3(1)--192C1(1) \\
                           & --\textit{N.Cl}.C1(1)--Gl.Avg.--Softmax \\
\multirow{1}{*}{Deeper}    & ~2$\times$96C3(1)--96C3(2)--2$\times$192C3(1)--192C3(2)--2$\times$192C3(1)\\
                           & --192C3(2)--192C3(1)--192C1(1)--\textit{N.Cl}.C1(1)--Gl.Avg.--Softmax \\
\end{tabular}
\end{center}
\end{table}

where $K$C$D$($S$) is a $D \times D$ convolutional layer with $K$ channels and stride $S$, followed by batch normalization and a ReLU non-linearity. \textit{N.Cl.} is the number of classes and Gl.Avg. refers to global average pooling. The network is identical to the All-CNN-C architecture in the original paper, except for the introduction of batch normalization. We set the same training parameters as in the original paper in the cases they are reported. Specifically, in all experiments the All-CNN networks are trained using stochastic gradient descent (SGD) with batch size of 128, during 350 epochs, with fixed momentum 0.9 and learning rate of 0.01 multiplied by 0.1 at epochs 200, 250 and 300. The kernel parameters are initialized according to the Xavier uniform initialization \cite{glorot2010glorot}.

\subsubsection{Wide Residual Network.}
WRN is a residual network \cite{he2016resnet} with more units per layer than the original ResNet, that achieves better performance with a smaller number of layers. In our experiments we use the WRN-28-10 version, with 28 layers and about 36.5 M parameters. The details of the architecture are the following:

\begin{center}
\centering
16C3(1)--4$\times$160R--4$\times$320R--4$\times$640R--BN--ReLU--Avg.(8)--FC--Softmax
\end{center}

where $K$R is a residual block with residual function  BN--ReLU--$K$C3(1)--BN--ReLU--$K$C3(1). BN is batch normalization, Avg.(8) is spatial average pooling of size 8 and FC is a fully connected layer. The stride of the first convolution within the residual blocks is 1 except in the first block of the series of 4, where it is 2 to subsample the feature maps. As before, we try to replicate the training parameters of the original paper: we use SGD with batch size of 128, during 200 epochs, with fixed Nesterov momentum 0.9 and learning rate of 0.1 multiplied by 0.2 at epochs 60, 120 and 160. The kernel parameters are initialized according to the He normal initialization \cite{he2015he}.

\subsection{Data}

We train the above described networks on both CIFAR-10 and CIFAR-100 \cite{krizhevsky2009cifar}. CIFAR-10 contains images of 10 different classes and CIFAR-100 of 100 classes. Both data sets consist of 60,000 32 x 32 color images split into 50,000 for training and 10,000 for testing. In all our experiments, the input images are fed into the network with pixel values in the range $[0, 1]$ and floating precision of 32 bits. Every network architecture is trained with three data augmentation schemes: no augmentation, light and heavier augmentation. The light scheme only performs horizontal flips and horizontal and vertical translations of 10\% of the image size, while the heavier scheme performs a larger range of affine transformations, as well as contrast and brightness adjustment. We use identical schemes as in  \cite{hernandez2018weightdecay_dropout}, where more details are given in an appendix. It is important to note though, that the light scheme is adopted from previous works such as \cite{goodfellow2013maxout,springenberg2014allcnn}, while the heavier scheme was first defined in \cite{hernandez2018weightdecay_dropout}, without aiming at designing a particularly successful scheme, but rather a scheme with a large range of transformations.

\subsection{Training and testing}

We train every model with the original explicit regularization, that is weight decay and dropout, as well as with no explicit regularization. Besides, we test both models with the three data augmentation schemes: light, heavier and no augmentation. The test accuracy we report results from averaging the softmax posteriors over 10 random \textit{light} augmentations.

All the experiments are performed on the neural networks API Keras \cite{chollet2015keras} on top of TensorFlow \cite{tensorflow2015_shorter} and on a single GPU NVIDIA GeForce GTX 1080 Ti.

\section{Results}

In this section we present and analyze the performance of the networks trained with different data augmentation schemes and with the regularizers on and off. We are interested in comparing data augmentation and explicit regularization regarding two different aspects: the performance when the training data set is reduced to 50~\% and 10~\% of the available examples and the performance when the architecture is shallower and deeper than the original. The presentation of the results in Figures~\ref{fig:lessdata} and~\ref{fig:depth} aims at enabling an easy comparison between the performance of a given network on a particular data set, when it has been trained with weight decay and dropout and when it has no explicit regularization (red and purple bars, respectively). The figures also allow a comparison of the performance between the different levels of regularization (color saturation).

\subsection{Reduced training sets}

\begin{figure}[h]
  \centering
  \begin{subfigure}{\linewidth}
      \includegraphics[keepaspectratio=true, width=\columnwidth]{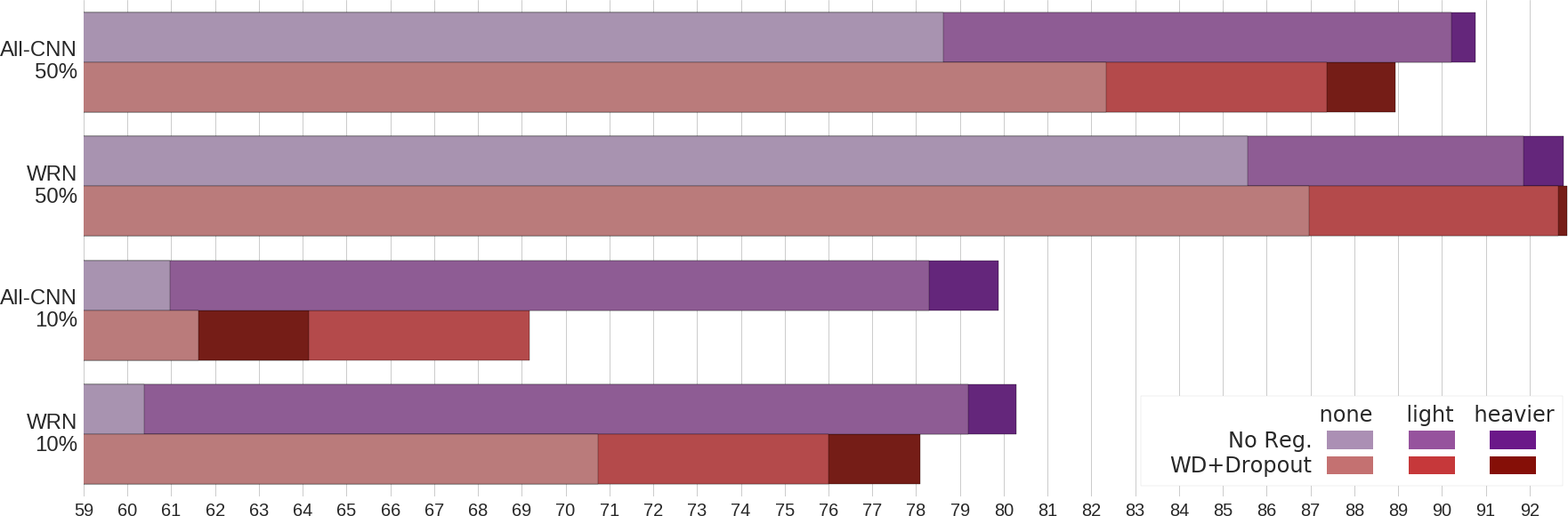}
      \caption{CIFAR-10}
    \label{fig:lessdata_cifar10}
  \end{subfigure}
  
  \begin{subfigure}{\linewidth}
      \includegraphics[keepaspectratio=true, width=\columnwidth]{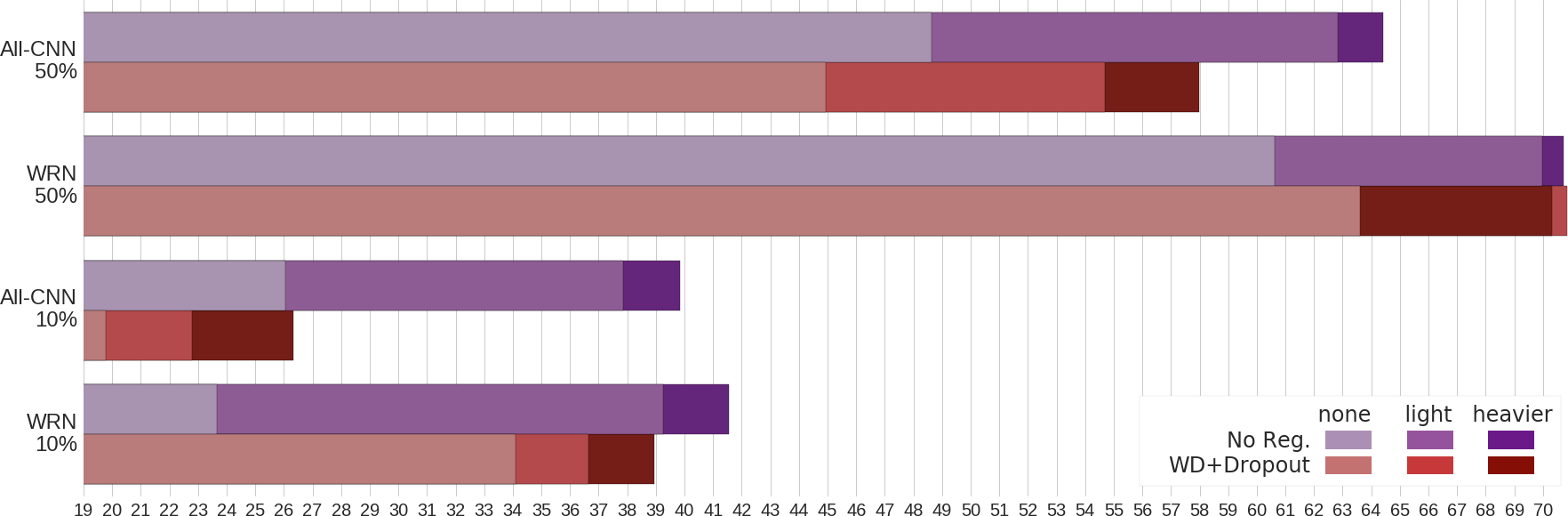}
      \caption{CIFAR-100}
      \label{fig:lessdata_cifar100}
  \end{subfigure}
  \caption{Test performance of the models trained with weight decay and dropout (red) and the models trained without explicit regularization (purple) when the amount of available training data is reduced. In general, the latter outperform the regularized counterparts and the differences become larger as the amount of training data decreases.}
  \label{fig:lessdata}
\end{figure}

The performance of All-CNN and WRN trained with only 50 and 10~\% of the available data is presented in Figure~\ref{fig:lessdata}. From a quick look at the accuracy bars it already becomes clear that the models trained without any explicit regularization (purple bars) outperform the models trained with weight decay and dropout (red bars). This is true for almost all the models trained with heavier data augmentation (darkest bars). Only in the case of WRN trained with 50~\% of CIFAR-10, the accuracy of the regularized model is marginally better ($< 0.001$). Otherwise, it seems that turning off the explicit regularizers not only does not degrade the performance, but it helps achieve even better generalization. 

The differences become even greater as the amount of training examples gets smaller, in view of the results of training with only 10~\% of the data. In these cases, the non-regularized models clearly outperform their counterparts. We hypothesize that this may occur because the value of the hyperparameters of weight decay and dropout, which were tuned to achieve state-of-the-art results with 100~\% of the data in the original publications, are not suitable anymore when the training data changes. It may be possible to improve the performance of the regularized models by adapting the value of the hyperparameters, but that would require a considerable amount of time and effort. On the contrary, it seems that the same data augmentation scheme helps generalize even when the training data set gets smaller.

The great implicit regularization effect of data augmentation becomes evident by looking at the large performance gap between the light scheme and no data augmentation. It seems that just a small set of simple transformations help the networks reduce the generalization gap by a large margin. In all cases the regularization effect is much larger than the one of weight decay and dropout.

\subsection{Shallower and deeper architectures}
\label{sec:depth}

\begin{figure}[ht]
  \centering
  \begin{subfigure}{\linewidth}
      \includegraphics[keepaspectratio=true, width=\columnwidth]{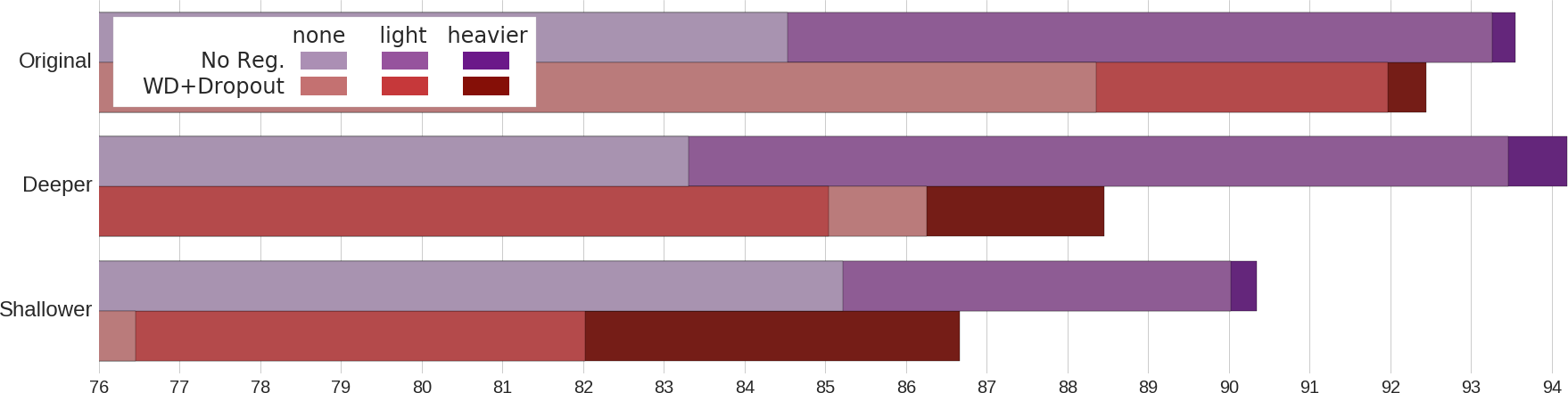}
      \caption{CIFAR-10}
    \label{fig:depth_cifar10}
  \end{subfigure}
  
  \begin{subfigure}{\linewidth}
      \includegraphics[keepaspectratio=true, width=\columnwidth]{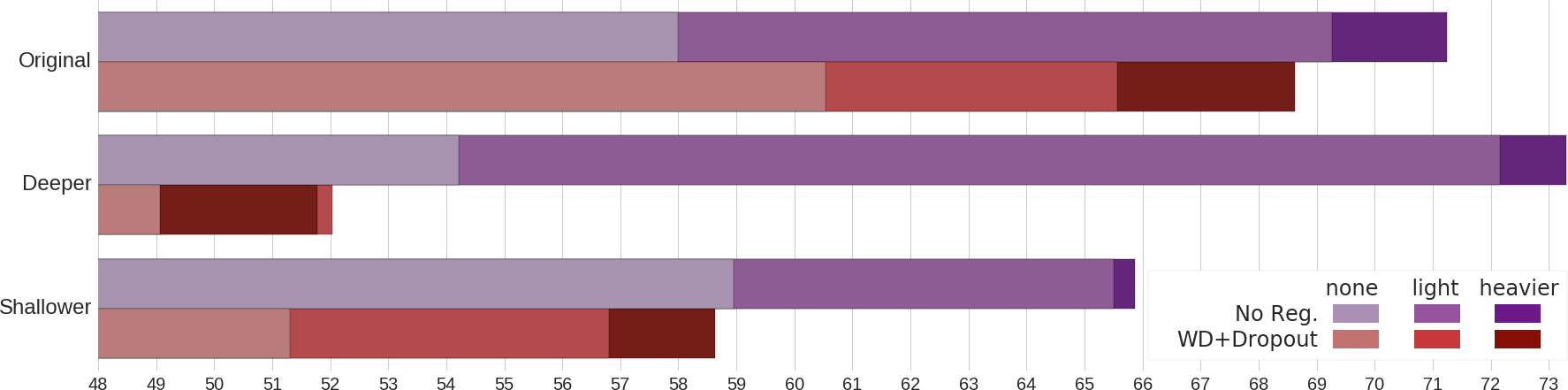}
      \caption{CIFAR-100}
      \label{fig:depth_cifar100}
  \end{subfigure}
  \caption{Test performance of the models trained with weight decay and dropout (red) and the models trained without explicit regularization (purple) on shallower and larger versions of All-CNN. In all the models trained with weight decay and dropout, the change of architecture results in a dramatic drop in the performance, compared to the models with no explicit regularization.}
  \label{fig:depth}
\end{figure}

Figure~\ref{fig:depth} shows the accuracy of All-CNN when we increase or reduce the depth of the architecture. If no explicit regularization is included (purple bars), we observe that the deeper architecture improves the results of the original network on both data sets, while the shallower architecture suffers a slight drop in the performance. In the case of the models with weight decay and dropout (red bars), not only is the performance much worse than their non-regularized counterparts, but even the deeper architectures suffer a dramatic performance drop. This seems to be another sign that the value of hyperparameters of weight decay and dropout largely depend on the architecture and any modification requires the fine-tuning of the regularization parameters. That is not the case of data augmentation, which again seems to easily adapt to the new architectures because its potential depends mostly on the type of training data.

\section{Discussion and Conclusion}

This work has extended the insights from \cite{hernandez2018weightdecay_dropout} about the futility of using weight decay and dropout for training convolutional neural networks for image object recognition, provided enough data augmentation is applied. In particular, we have focused on further exploring the advantages of data augmentation over explicit regularization, in terms of its adaptability to changes in the network architecture and the size of the training set. 

Our results show that explicit regularizers, such as weight decay and dropout, cause significant drops in performance when the size of the training set or the architecture changes. We believe that this is due to the fact that their hyperparameters are highly fine-tuned to some particular settings and are extremely sensitive to variations of the initial conditions. On the contrary, data augmentation adapts more naturally to the new conditions because its hyperparameters, that is the type of transformations, depend on the type of training data and not on the architecture or the amount of available data. For example, a model without neither weight nor dropout slightly improves its performance when more layers are added and therefore the capacity is increased. However, with explicit regularization, the performance even decreases.

%
These findings contrast with the standard practice in the convolutional networks literature, where the use of weight decay and dropout is almost ubiquitous and believed to be necessary for enabling generalization. Furthermore, data augmentation is sometimes regarded as a hack that should be avoided in order to test the potential of a newly proposed architecture. We believe instead that these roles should be switched, because in addition to the results presented here, data augmentation has a number of other advantages: it increases the robustness of the models against input variability without reducing the effective capacity and may also enable learning more biologically plausible features \cite{hernandez2018dataaugit}. We encourage future work to shed more light on the benefits of data augmentation and the handicaps of ubiquitously using explicit regularization, specially on research projects, by testing new architectures and data sets.

\subsection*{Acknowledgments}

This project has received funding from the European Union's Horizon 2020 research and innovation programme under the Marie Sklodowska-Curie grant agreement No 641805.

\bibliographystyle{iclr2018_conference}
\bibliography{references}

\end{document}